\documentclass[11pt]{article}

\usepackage[final]{acl}

\usepackage{times}
\usepackage{latexsym}
\usepackage[T1]{fontenc}
\usepackage[utf8]{inputenc}
\usepackage{microtype}
\usepackage{inconsolata}
\usepackage{graphicx}
\usepackage{amsmath}
\usepackage{amssymb}
\usepackage{booktabs}
\usepackage{multirow}
\usepackage{xcolor}
\usepackage{colortbl}
\usepackage{enumitem}
\usepackage{algorithm}
\usepackage{algpseudocode}
\usepackage{subcaption}
\usepackage{tikz}
\usepackage[T1]{fontenc}
\usepackage{newtxtt}
\usetikzlibrary{positioning, arrows.meta, shapes.geometric, calc, fit, backgrounds}

\definecolor{lightgray}{gray}{0.92}

\newcommand{\ours}{VISA}

\title{\ours{}: Agentic Self-Evolving Data Synthesis for Multimodal Instruction Following}

\author{
\textbf{Min Zeng},
\textbf{Guanxin Tan},
\textbf{Libin Cen},
\textbf{Yafei Wen} \\
\textbf{Rui Hu},
\textbf{Liuyang Bian},
\textbf{Xiaolong Chen},
\textbf{Xiaoxin Chen} \\
vivo AI Lab \\
\small{\href{mailto:zengmin325@163.com}{zengmin325@163.com}}
}

\begin{document}
	\maketitle
	
	\begin{abstract}
    Multimodal instruction-following models require training data that is accurate, diverse, verifiable, and challenging. Existing synthesis pipelines typically follow a one-pass generate-and-filter paradigm, discarding feedback from failed samples, verifier outcomes, and target-model errors. We present \textsc{VISA} (Visual Instruction Synthesis Agent), an agentic framework that reformulates multimodal instruction synthesis as a self-evolving loop. At each round, \textsc{VISA} analyzes an image to filter incompatible constraints and discover new verifiable ones, samples diversity- and difficulty-aware constraint sets from persistent memory, generates candidate instructions, and verifies the resulting samples with executable tools and structured large language model judges. Failed samples trigger diagnostic-guided recovery, while accepted samples are probed against the target model to estimate difficulty. The resulting verifier signals and target-model failure profiles are written back to memory, allowing subsequent rounds to adaptively expand the constraint space, reduce template repetition, and focus on unresolved model weaknesses. The same verifier contracts further provide reward signals for reinforcement learning without a separately trained reward model. Experiments on MM-IFEval show that \textsc{VISA} consistently improves multimodal instruction following over strong baselines, while preserving general multimodal capability across seven public benchmarks.
	\end{abstract}

	\section{Introduction}
	\label{sec:intro}
	
	Multimodal large language models (MLLMs) have achieved remarkable progress in visual understanding and reasoning abilities \citep{liu2023visual, team2026qwen3, bai2025qwen3, li2024llava}. Beyond perception, real-world applications increasingly require MLLMs to precisely follow complex user instructions under diverse constraints, such as output formatting, keyword inclusion, style control, structured generation, and visually grounded reasoning \citep{zhou2023instruction, qian2025mia}. Building MLLMs that can reliably satisfy such multimodal constraints has therefore become an important problem for multimodal alignment and instruction tuning.
	
	Recent works improve multimodal instruction following by synthesizing constraint-rich data and designing specialized evaluation benchmarks \citep{ding2025mm, he2026empowering}. However, most synthesis pipelines are still one-pass systems. They generate instructions from templates or prompts, filter the outputs, and then add the accepted samples to a training set. Feedback from failed samples, repeated templates, missing verifiers, or target-model errors is often used only locally, if at all. As a result, the synthesis process lacks an agent loop: it does not reflect on its past outputs, update its internal state, or use that state to plan.
	
	In this paper, we present VISA (Visual Instruction Synthesis Agent), a self-evolving agentic framework that treats multimodal instruction synthesis as a closed-loop optimization problem rather than a one-shot generation task. At its core, VISA runs an agent loop between a planning and execution stage, which composes image-grounded instructions and generates responses, and a reflection stage, which evaluates each sample through a unified dual-track verifier combining executable code tools and structured large language model (LLM) judges. When a sample fails verification, the loop triggers diagnostic-guided recovery; when it passes, the loop probes the target model to estimate difficulty. Crucially, the outcomes of every iteration—accepted samples, constraint–verifier bindings, diversity statistics, and per-constraint failure rates—are written into a persistent memory that reshapes the next round's constraint sampling and instruction planning. Through this iterative accumulation, VISA continuously steers synthesis toward data that is more accurate, more diverse, and more challenging to the current target model, without additional human annotation during self-evolution.
	
	Our contributions are three-fold: (1) We propose a self-evolving framework for multimodal instruction data synthesis that turns static generation into an iterative feedback-driven process; (2) We introduce a unified verification and recovery framework that supports dynamic verifier binding, bounded agentic repair, and target-aware difficulty probing during synthesis, while naturally producing verifiable reward signals that can be readily extended to reinforcement learning (RL); (3) We demonstrate through extensive experiments on multiple multimodal instruction-following benchmarks that VISA consistently improves instruction adherence while preserving strong general multimodal capabilities.
	
	\begin{figure*}[t]
	\centering
	\includegraphics[width=1.95\columnwidth]{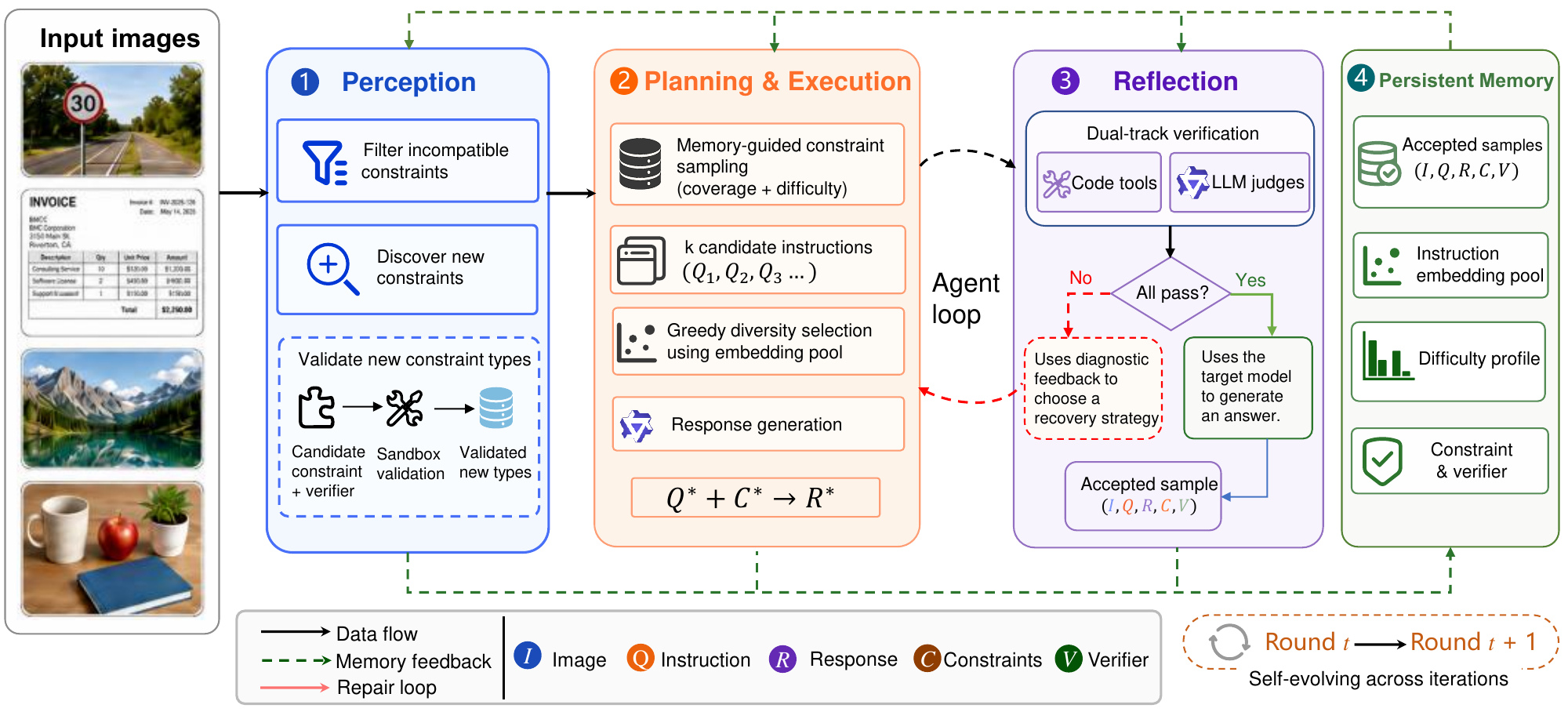}
	\caption{Overview of VISA. The framework iteratively performs perception, planning and execution, reflection, and memory update, allowing verifier feedback and target-model difficulty signals to guide subsequent synthesis rounds.}
	\label{fig:framework}
\end{figure*}

	\section{Related Work}
	
	Existing multimodal instruction synthesis pipelines primarily scale data diversity and response quality through large-scale synthetic generation~\citep{huang2025mm, zhang2024multimodal}, with recent works such as MM-IFEngine~\citep{ding2025mm} further introducing structured constraints and hybrid verification to improve instruction-following quality. Self-evolving approaches such as Self-Instruct~\citep{wang2023self}, WizardLM~\citep{xu2023wizardlm}, and MMEvol~\citep{luo2025mmevol} improve instruction complexity through iterative rewriting or evolutionary generation. However, all of these methods remain fundamentally open-loop: they optimize what is generated but leave the synthesis process itself static, with no mechanism to adapt based on what the target model has yet to master. VISA instead treats synthesis as a closed-loop optimization process, where verification outcomes, difficulty signals, and accumulated memory continuously reshape subsequent rounds of data generation.
	
	Prior works on instruction-following evaluation~\citep{zhou2023instruction, qian2025mia} demonstrate that many constraints admit precise programmatic verification—an insight VISA extends from evaluation into synthesis by binding each constraint to a verifier at generation time. Agentic LLM systems~\citep{yao2022react, hong2024metagpt, qin2024toolllm} similarly decompose tasks into perception, planning, and execution stages, but typically operate within isolated episodes without persistent state. VISA maintains a persistent memory across synthesis rounds, so that reflection outcomes directly reshape future planning rather than being discarded at the end of each episode.
	
	\section{Method}
	\label{sec:method}	
	VISA casts multimodal instruction synthesis as a closed-loop process with four stages: Perception, Planning \& Execution, Reflection, and Memory Update. As shown in Figure~\ref{fig:framework}, each round generates, verifies, and repairs candidate samples, then writes accepted samples and difficulty signals into persistent memory to guide the next round.
    
	\subsection{Problem Formulation}
	\label{sec:problem}
	
	We frame synthesis as image-conditioned constraint satisfaction.
	Given an image pool $\mathcal{I}$ and a dynamically evolving constraint-type registry $\mathcal{T}$, VISA constructs a multimodal instruction-following dataset
	\begin{equation}
		\mathcal{D} = \{(I_i,\, Q_i,\, R_i,\, C_i,\, v_i)\}_{i=1}^{N},
	\end{equation}
	where $I_i \in \mathcal{I}$ is an input image, $Q_i$ is an instruction grounded in $I_i$, $C_i \subset \mathcal{T}$ is the set of constraints embedded in $Q_i$, $R_i$ is the corresponding response expected to satisfy every $c \in C_i$, and $v_i$ denotes the verification procedure—either a code-based verifier or a structured LLM judge—bound to the sample at synthesis time.
	The registry $\mathcal{T}$ is not fixed: new constraint types may be proposed and validated during Perception, allowing the synthesized data to expand beyond its initial scope without human annotation.
	
	\subsection{Perception}
	\label{sec:perception}
	
	A fundamental limitation of static synthesis pipelines is that a hand-crafted constraint registry, however carefully designed, cannot anticipate every instruction requirement that naturally arises from a diverse image collection. The Perception stage addresses this by first analyzing each image with the synthesis model to identify which constraint types in $\mathcal{T}$ are meaningful for the current scene, returning a compatible subset $\mathcal{S}_I \subseteq \mathcal{T}$ by pruning types whose requirements cannot be grounded in the visual content—a \texttt{spatial\_relation} constraint, for instance, is semantically vacuous for a close-up portrait. Restricting downstream planning to $\mathcal{S}_I$ ensures that every selected constraint is both verifiable and visually meaningful, preventing the planner from composing instructions that are technically valid but disconnected from the image.
	
	Beyond filtering, the Perception stage actively encourages the synthesis model to propose constraint types that are uniquely suited to each image, even when the current registry already provides adequate coverage.
	This image-driven discovery enriches the constraint space with fine-grained, visually grounded requirements that a static registry would not express, directly improving the diversity of synthesized instructions.
	Each proposed type is paired with a candidate verification function and must pass a sandboxed validation test before admission to $\mathcal{T}$, ensuring that every newly discovered constraint remains reliably verifiable in subsequent iterations.
	
	\subsection{Planning \& Execution}
	\label{sec:planexec}
	Given the image-compatible constraint subset $\mathcal{S}_I$, a naive planner would repeatedly sample similar constraint combinations and generate near-duplicate instructions across iterations—a form of data homogenization that limits the diversity and informativeness of the training corpus. VISA addresses this by jointly managing diversity at two levels: the constraint level, guided by memory signals, and the instruction level, guided by embedding-space distance. Specifically, constraint sampling weights each type in $\mathcal{S}_I$ by two signals maintained in persistent memory—a coverage signal that down-weights types already well-represented in the accepted corpus, and a difficulty signal that up-weights types whose constraints the target model has historically struggled to satisfy (derived from the difficulty probing described in reflection)—concentrating synthesis on combinations that are simultaneously underrepresented and challenging.
	
	To alleviate data homogenization, VISA separates instruction planning from response generation. For each image, the planner first generates $k$ candidate instructions with sampled constraint sets, without immediately producing the corresponding responses. Each candidate instruction is then embedded and compared with the global embedding pool $\mathcal{P}$ of previously accepted instructions stored in memory. Following a greedy diversity selection strategy, VISA selects the candidate whose nearest neighbor in $\mathcal{P}$ is farthest away:
	\[
	i^{*} = \arg\max_{i \in [k]} \min_{p \in \mathcal{P}} \mathrm{dist}(e_{Q_i}, e_p),
	\]
	where $e_{Q_i}$ denotes the embedding of candidate instruction $Q_i$, $e_p$ denotes an instruction embedding in $\mathcal{P}$, and $\mathrm{dist}(\cdot,\cdot)$ measures the distance between two embeddings. Only the selected instruction $Q_{i^{*}}$ is passed to the response generator to produce $R_i$. 
		
	\subsection{Reflection}
	\label{sec:reflection}
	
	After planning and execution, the Reflection stage determines whether a generated sample should be accepted, repaired, or discarded. Given a sample $(I, Q, R, C)$, VISA evaluates each constraint $c \in C$ through its bound verifier: deterministic constraints—word count, keyword inclusion, output format—are checked by executable code tools, while semantic and visually grounded constraints are assessed by structured LLM judges, both sharing a unified output interface that reports a binary pass/fail verdict alongside diagnostic information localizing each failure. When a sample fails, VISA uses these per-constraint diagnostics as structured feedback to select the most appropriate recovery strategy—applying local response edits for surface-level violations, or escalating to instruction rewrites and constraint simplification when the failure reflects a deeper incompatibility between the instruction and the image—rather than discarding the sample outright. If the verification score stops improving across consecutive attempts, VISA discards the sample once the budget is exhausted, preventing the degenerate behavior in which repeated self-correction converges to an incorrect solution.
	
	For samples that pass verification, VISA estimates their informativeness by probing the target model $M$: given the accepted instruction $Q$ and image $I$, $M$ generates a response that is then evaluated against the same constraint set used during synthesis. The per-constraint outcomes are aggregated into a difficulty score $d \in [0, 1]$, where a lower score indicates that $M$ already satisfies most constraints and thus gains little from the sample, while a higher score signals unresolved failures that provide more informative supervision. Both the aggregate score and the per-constraint failure pattern are written to memory, subsequently shifting the constraint sampling weights in the Planning stage toward types that remain challenging for $M$ and directly linking each synthesis round to the current capability profile of the target model.
	
	\subsection{Memory}
	\label{sec:memory}
	
	A defining characteristic of most data synthesis pipelines is that they are stateless: each sample is generated independently, and the synthesis process itself does not improve as more data is produced. VISA breaks this assumption by maintaining a persistent memory $\mathcal{M}$ that accumulates the outcomes of every synthesis iteration and makes them available to all subsequent rounds. Specifically, $\mathcal{M}$ stores the accepted samples together with their constraint sets and verifier handles, the instruction embeddings that populate the diversity pool $\mathcal{P}$, the per-constraint difficulty profiles derived from target-model probing, and the constraint types discovered during Perception along with their associated verification functions.
	
	This accumulated state feeds directly back into every stage of the next iteration, giving rise to VISA's self-evolving behavior. The instruction embedding pool steers the diversity selector away from redundant instructions; the difficulty profiles shift constraint sampling weights toward types that remain unresolved for the target model; and newly discovered constraint types become immediately available for future synthesis. As a result, each iteration begins with strictly more information than the last—about what has already been generated, where the target model still struggles, and what constraints the system can express—so that subsequent rounds can exploit increasingly informative synthesis state without external intervention.

	\section{Reinforcement Learning with Verifiable Rewards}
	
	\label{sec:rl}
	
	Recent studies on reinforcement learning with verifiable rewards (RLVR) show that rule-based or programmatically checkable signals can guide policy optimization without relying on separately trained reward models \citep{wen2025reinforcement, cai2025reinforcement}. This idea has also been adopted in recent reasoning models such as DeepSeek-R1 \citep{guo2025deepseek}, where Group Relative Policy Optimization (GRPO) optimizes model responses using verifiable reward signals for tasks with clearly defined correctness criteria \citep{shao2024deepseekmath}. These methods suggest that verifier-derived rewards can provide a more transparent and task-aligned alternative to separately learned reward models when the desired behavior can be checked by explicit rules or reliable evaluators.
	
	VISA naturally fits this paradigm because the agent binds each accepted instruction to explicit verifier contracts during execution. The same verification logic used to accept and repair synthetic data can therefore be reused during RL. Given an image $I$, an instruction $Q$, a model response $R$, and the constraint set $C=\{c_1,\ldots,c_m\}$ embedded in the instruction, each constraint $c_j$ is evaluated by its bound verifier $v_j$. We let $v_j(I,Q,R)\in\{0,1\}$ denote whether the response satisfies $c_j$, and define the reward as the constraint pass rate:
	\[
	r(I,Q,R,C)=\frac{1}{|C|}\sum_{j=1}^{|C|} v_j(I,Q,R).
	\]
	This reward directly measures the fraction of instruction constraints followed by the model response, aligning RL optimization with multimodal instruction following. Since the reward is computed from verifier contracts already maintained by the synthesis agent, VISA does not require an additional reward model, and failed constraints can be traced to specific sources such as formatting errors, missing keywords, or incorrect visually grounded descriptions.

	\section{Experimental Setup}
	\label{sec:experiments}

	\subsection{Generated Training Data}
	\label{sec:generated_data}
	We synthesize a dataset of approximately 15k samples using VISA, based on the image pool provided by MM-IFInstruct\footnote{\url{https://huggingface.co/datasets/ChrisDing1105/MMIF-23k}}. Each generated sample includes an image, a corresponding instruction, the model's response, and the associated verification method. This dataset is used for fine-tuning of multimodal instruction-following models in our experiments.
	
	\subsection{Models}
	\label{sec:exp_model}
    We use the Qwen3.5\footnote{\url{https://huggingface.co/collections/Qwen/qwen35}} series as the primary model family in our experiments. Specifically, Qwen3.5-27B serves as the synthesis agent in VISA, while Qwen3.5-4B is used as the target model for supervised fine-tuning (SFT) and RL. We also include Qwen3.5-9B as a stronger same-family reference model. For comparison with contemporary open-source MLLMs, we report results of MiniCPM-V-4.5 \citep{yu2025minicpm} and InternVL3.5 \citep{wang2025internvl3} at comparable model scales.
	
	\subsection{Training Details}
	\label{sec:training_details}
	All experiments are conducted on a cluster of 128 L40s GPUs. The visual encoder layers are frozen, and only the LLM backbone and the alignment module are fine-tuned. We adopt a learning rate of $1\times10^{-5}$ and perform training in bfloat16 precision to balance memory efficiency and numerical stability. To efficiently scale across multiple GPUs, we utilize DeepSpeed~\citep{rajbhandari2020zero} ZeRO-3 for parallelized training, enabling effective model updates across the large distributed setup. SFT is implemented using swift~\citep{zhao2025swift}, while RL stages leverage verl~\citep{sheng2024hybridflow}, with rollout size set to 8 and GPU memory utilization at 0.8.
	
	\subsection{Evaluation Benchmarks}
	\label{sec:eval_benchmarks}
	
	We evaluate VISA from two perspectives: multimodal instruction following and general multimodal capability. For instruction following, we use MM-IFEval~\citep{ding2025mm}, which evaluates both compose-level (C-Level) constraints on output responses and perception-level (P-Level) cases grounded in input images. We report C-Level, P-Level, and average accuracy following the original protocol. For general capability evaluation, we report results on MMBench~\citep{liu2024mmbench}, MMStar~\citep{chen2024we}, MM-Vet~\citep{yu2023mm}, HallusionBench~\citep{guan2024hallusionbench}, MathVista~\citep{lu2024mathvista}, OCRBench~\citep{liu2024ocrbench}, and AI2D~\citep{kembhavi2016diagram}. These benchmarks cover broad multimodal perception and reasoning, integrated vision-language abilities, hallucination robustness, visual mathematical reasoning, optical character recognition (OCR)-centric understanding and scientific diagram understanding. To ensure a fair comparison with the base model and avoid confounding improvements from explicit reasoning traces, all benchmark evaluations are conducted in the non-thinking mode.


	\section{Results}
	\label{sec:results}

	\begin{table}[t]
	\centering
	\small
	\setlength{\tabcolsep}{3.2pt}
	\renewcommand{\arraystretch}{1.08}
	\begin{tabular}{@{}lc|ccc@{}}
		\toprule
		\multirow{2}{*}{\textbf{Model / Training}} &
		\multirow{2}{*}{\textbf{Params}} &
		\multicolumn{3}{c}{\textbf{MM-IFEval}} \\
		& & C-Level & P-Level & Avg. \\
		\midrule
		Qwen3.5-9B & 9B & 63.9 & 60.0 & 63.0 \\
        MiniCPM-V-4.5& 8B & 68.8 & 44.0 & 62.6 \\
		InternVL3.5-8B & 8B & 61.8 & 39.0 & 56.1 \\
		InternVL3.5-4B & 4B & 57.0 & 33.0 & 51.0 \\
        Qwen3.5-4B & 4B & 62.7 & 55.0 & 60.8 \\
		\midrule
		Qwen3.5-4B\\
		\quad + MM-IFInstruct-23k & 4B & 65.7 & 46.0 & 60.8 \\
		\quad + MM-IFDPO-23k & 4B &  65.0 &  52.0 &  61.7 \\
		\quad + \textbf{\ours{}-SFT-15k} & 4B & 67.2 & 54.0 & 63.9 \\
		\quad + \textbf{\ours{}-RL-15k} & 4B & 66.9 & 59.0 & 64.9 \\
		\bottomrule
	\end{tabular}
	\caption{Main results on MM-IFEval. We report C-Level accuracy, P-Level accuracy, and their average score; higher values indicate better instruction-following performance.}
	\label{tab:main}
\end{table}

\subsection{Main Results}

Table~\ref{tab:main} presents the main results on MM-IFEval. Among the baselines, Qwen3.5-9B achieves an average score of 63.0, while the 4B-scale models (MiniCPM-V-4.5, InternVL3.5-4B, and Qwen3.5-4B) score 62.6, 51.0, and 60.8, respectively. Training Qwen3.5-4B on MM-IFInstruct-23k brings only limited improvement, suggesting that simply increasing data volume without targeted synthesis is insufficient. In contrast, VISA-SFT-15k achieves an average score of 63.9, surpassing the 9B reference model on the average score with less than half the parameters and using only 15k training samples. VISA-RL-15k further improves the average score to 64.9, with a notable gain on P-Level (59.0), reducing the performance gap between constraint-oriented and perception-oriented cases.

MM-IFEval contains 300 C-Level samples that assess multi-constraint satisfaction in model outputs, and 100 P-Level samples that require visually grounded understanding akin to visual question answering (VQA), where instruction-following ability is less likely to be the only bottleneck. As a result, P-Level gains are less directly attributable to instruction-following supervision alone, since they also depend on the model's underlying visual perception and grounding ability. Nevertheless, VISA-RL-15k attains a P-Level score of 59.0, outperforming all 4B-scale baselines and approaching Qwen3.5-9B (60.0), suggesting that difficulty-aware constraint sampling and verifier-based RL provide useful supervision even for perception-heavy instructions.

	\label{sec:main_results}
\begin{table}[t]
\centering
\small
\begin{tabular}{lccc}
	\toprule
	Method (Qwen3.5-4B) & C-Level & P-Level & Avg. \\
	\midrule
	Base							  & 62.7 & 55.0 & 60.8 \\
	Static Pipeline                   & 64.9 & 46.0 & 60.2 \\
	\quad + Reflection                & 65.6 & 48.0 & 61.2 \\
	\quad + Reflection + Memory       & 67.1 & 51.0 & 63.1 \\
	Full VISA                         & 67.2 & 54.0 & 63.9 \\
	\bottomrule
\end{tabular}
\caption{Ablation results on MM-IFEval with Qwen3.5-4B under SFT, showing the incremental contribution of reflection, memory, and the full VISA pipeline.}
\label{tab:ablation}
\end{table}

	\subsection{Ablation Study}
	\label{sec:ablation}
    To understand the contribution of each component in VISA, we conduct an ablation study on MM-IFEval using Qwen3.5-4B as the target backbone under the SFT setting. As shown in Table~\ref{tab:ablation}, the static pipeline improves C-Level but noticeably degrades P-Level, leading to a lower average score than the base model. This suggests that naively generating constraint-rich data can improve output-constraint adherence while harming perception-oriented behavior. Adding the Reflection module, which performs diagnostic-guided repair of failed samples, improves the average score from 60.2 to 61.2, indicating that bounded recovery enhances per-sample reliability. Incorporating persistent memory further raises the average score to 63.1, with consistent gains on both C-Level and P-Level, confirming that coverage and difficulty signals accumulated across rounds help steer synthesis toward underrepresented and challenging constraints. The full VISA system achieves the best performance, reaching 67.2 on C-Level, 54.0 on P-Level, and 63.9 on average. Compared with the static pipeline, the largest numerical recovery appears on P-Level, suggesting that reflection and memory help reduce the perception-side degradation introduced by naive synthesis. However, since P-Level cases are more perception-oriented and closer to VQA-style evaluation, the C-Level gains provide more direct evidence that VISA improves compositional instruction following. These results show that the components contribute complementarily: Reflection improves local sample quality, Memory guides global constraint selection, and the remaining modules further enhance instruction diversity and difficulty targeting.
    
\begin{table*}[t]
	\centering
	\small
	\setlength{\tabcolsep}{2.2pt}
	\renewcommand{\arraystretch}{1.08}
	\begin{tabular}{@{}lccccccccc@{}}
		\toprule
		\textbf{Model / Training}
		 & MMBench & MMStar & MM-Vet	& HallusionBench	& MathVista & OCRBench & AI2D & Avg. \\
		\midrule
		Qwen3.5-4B   & 78.4 & 61.7 & 47.6 & 56.0 & 78.8 & 84.9 & 86.2 &70.5 \\
		\quad + \textbf{\ours{}-SFT-15k} & 81.2 & 63.2 & 49.1 & 54.1 & 77.8 & 85.2 & 86.4 & 71.0 \\
		\quad + \textbf{\ours{}-RL-15k}   & 78.8 & 62.5 & 50.5 & 57.5 & 87.9 & 85.6 & 87.6 & 72.9\\
		\bottomrule
	\end{tabular}
	\caption{General multimodal capability results on Qwen3.5-4B. VISA preserves or improves broad multimodal performance across seven public benchmarks, with all evaluations conducted in non-thinking mode.}
	\label{tab:general}
\end{table*}
    
\subsection{General Capability Evaluation}
\label{sec:general}

Table~\ref{tab:general} reports results across seven general multimodal benchmarks. VISA-SFT-15k improves the average score from 70.5 to 71.0, with gains on MMBench ($+$2.8) and MMStar ($+$1.5), and only a minor drop on HallusionBench ($-$1.9). These results suggest that instruction tuning on VISA data broadly preserves, and in some cases slightly enhances, general multimodal perception and reasoning. VISA-RL-15k achieves the strongest overall performance with an average score of 72.9, improving over the base model on five out of seven benchmarks. The largest gain appears on MathVista ($+$9.1), which may partly reflect the alignment between verifier-derived optimization and structured visual reasoning tasks, as MathVista emphasizes mathematical reasoning in visual contexts. OCRBench and AI2D also improve under RL, while HallusionBench also recovers to 57.5 compared with the SFT variant, suggesting that RL does not exacerbate the mild hallucination-related drop observed after SFT. Together, these results show that VISA improves instruction adherence without sacrificing general multimodal capability, and that the verifier-derived reward signal supports transferable improvements across diverse evaluation settings.

		\begin{figure*}[t]
		\centering
		\begin{subfigure}[b]{0.49\textwidth}
			\centering
			\includegraphics[width=\linewidth]{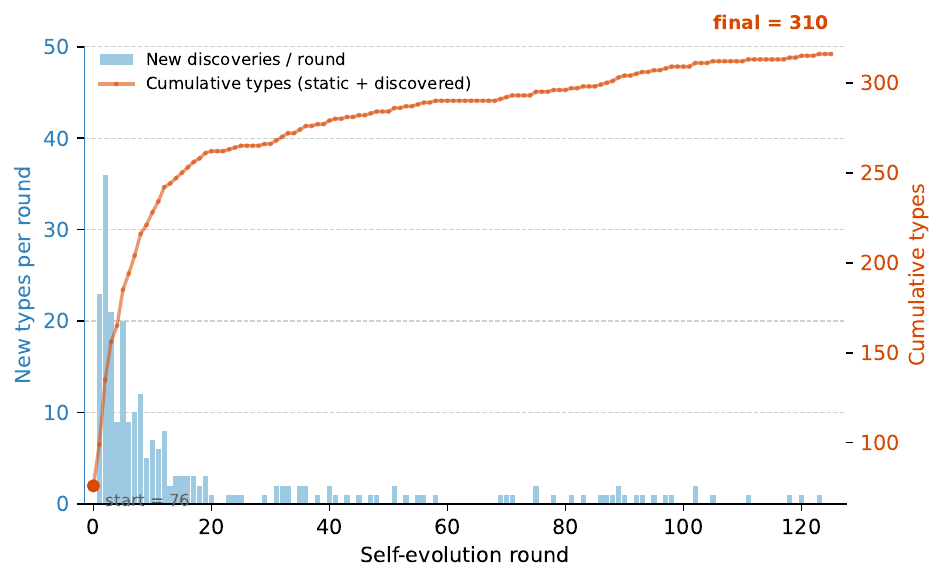}
			\label{fig:constraint_evolution}
		\end{subfigure}
		\hfill
		\begin{subfigure}[b]{0.49\textwidth}
			\centering
			\includegraphics[width=\linewidth]{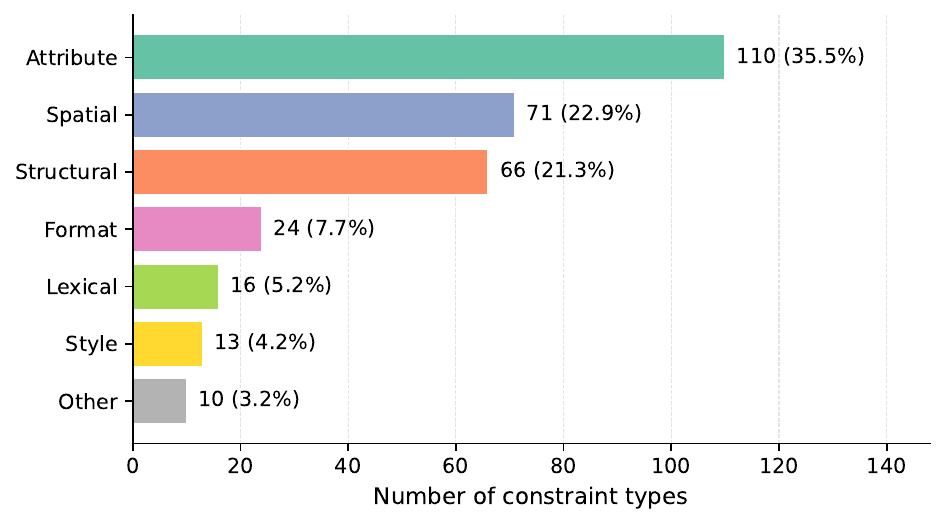}
			\label{fig:constraint_distribution}
		\end{subfigure}
		\caption{Evolution of the constraint-verifier registry in VISA. Left: cumulative growth of constraint types across self-evolution rounds, showing that most discoveries occur early and gradually saturate. Right: category distribution of the final 310 constraint types.}
		\label{fig:constraint_overview}
	\end{figure*}

	\section{Analysis}
	\label{sec:analysis}
	
	\subsection{Dynamic Constraint Expansion}
	\label{sec:mmifeval_breakdown}

	Figure~\ref{fig:constraint_overview} analyzes how the agent's constraint-verifier registry grows during data synthesis. We use \emph{constraint category} to denote a high-level group such as \texttt{attribute}, \texttt{spatial}, or \texttt{format}, and \emph{constraint type} to denote a fine-grained requirement within a category, such as \texttt{word\_count}, \texttt{json\_format}, or \texttt{spatial\_relation}. Starting from $76$ hand-crafted constraint types, \textsc{VISA} expands the registry during perception. For each image, the system analyzes what kinds of requirements the image can support and may propose new constraint types when the current state does not cover a useful image-grounded requirement. Each new type is linked to either a deterministic code verifier or a structured LLM judge before it is added to the registry. Across $126$ self-evolution rounds, this process adds $234$ new constraint types, producing a final library of $310$ types, a $4.1\times$ increase without additional human annotation. Appendix~\ref{app:constraint_details} lists the corresponding constraint categories and representative types.

	The left panel of Figure~\ref{fig:constraint_overview} shows that most new constraint types are found in the early rounds. After this initial stage, the number of newly added types drops quickly, and the cumulative curve becomes almost flat. This pattern suggests that the agent first discovers common visual requirements, such as object attributes, spatial relations, and response structure, and then only occasionally adds rare cases in later rounds. In other words, the registry grows quickly at the beginning but does not keep expanding without limit. The right panel shows that the final registry is mainly composed of \emph{attribute}, \emph{spatial}, and \emph{structural} categories. This means that the newly added types are mostly related to visual content and response organization, rather than simple text-only rules such as keyword matching or output format.

	\subsection{Diversity and Difficulty}
	\label{subsec:diversity}
	
	\begin{figure*}[t]
		\centering
		\begin{subfigure}[t]{0.46\textwidth}
			\centering
			\includegraphics[width=\linewidth]{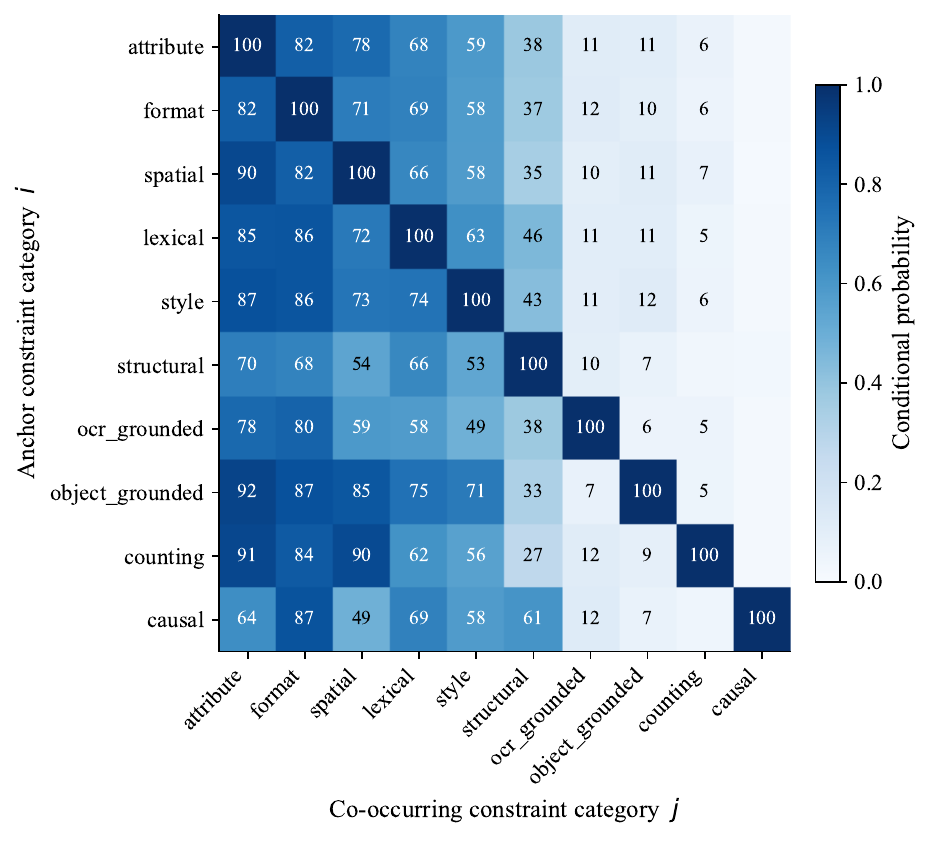}
		\end{subfigure}
		\hfill
		\begin{subfigure}[t]{0.46\textwidth}
			\centering
			\includegraphics[width=\linewidth]{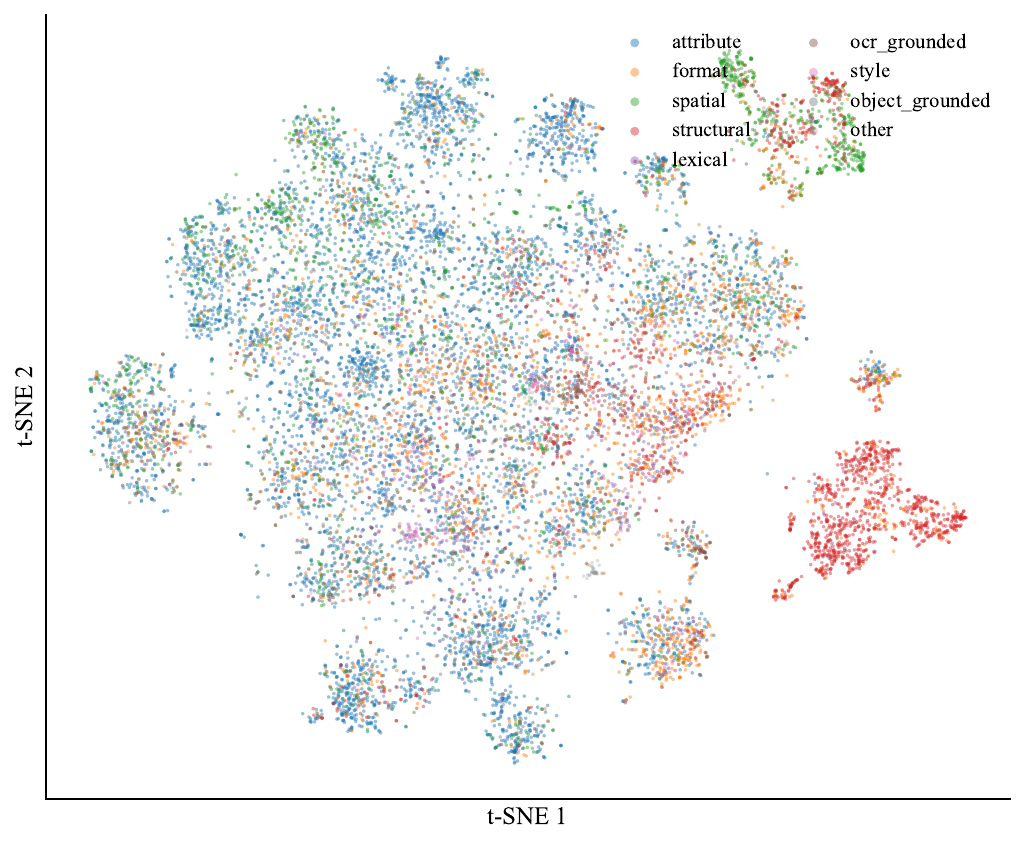}
		\end{subfigure}
		\caption{Diversity analysis of VISA-synthesized instructions. Left: conditional co-occurrence among the ten most frequent constraint categories, where each cell reports $P(j\mid i)$ as a percentage. Right: t-distributed stochastic neighbor embedding (t-SNE) projection of high-dimensional instruction embeddings, colored by dominant constraint category.}
        
		\label{fig:diversity_combo}
	\end{figure*}
	
	We analyze the $15{,}668$ samples produced by \textsc{VISA} from the perspectives of diversity and difficulty. Figure~\ref{fig:diversity_combo} shows that the data covers a broad joint constraint space. All $90$ off-diagonal category pairs appear at least once, and the strongest co-occurrences follow intuitive visual dependencies: \texttt{object\_grounded}, \texttt{counting}, and \texttt{spatial} co-occur with \texttt{attribute} in $85$--$92\%$ of samples, while \texttt{causal} pairs more often with \texttt{format} ($87\%$) and \texttt{structural} ($61\%$) than with \texttt{spatial} ($49\%$). The instruction embedding view further shows a continuous cloud rather than a few isolated template clusters, with a mean pairwise cosine distance of $0.977$. This supports the role of the embedding pool in planning: new instructions are selected against previous accepted instructions, reducing repeated surface forms. The full constraint-count distribution is provided in Figure~\ref{fig:constraint_count} in Appendix~\ref{app:additional_analysis}; it is centered around four to six constraints per sample (mean $4.87$, median $5$, standard deviation $2.12$), confirming that most examples combine several jointly satisfiable requirements instead of testing a single rule.
	
    We also use the target-model probe to estimate how much training signal these samples contain. Across all samples, $82.6\%$ are labeled \emph{easy}, $15.9\%$ \emph{medium}, and $1.5\%$ \emph{hard}, with $5.6\%$ of constraint instances flagged as failures by the verifier track. This skewed distribution is expected, since the target model already possesses strong basic multimodal instruction-following ability, making many verified samples easy under the probing protocol. Rather than enforcing a balanced sample-level difficulty distribution, \textsc{VISA} uses difficulty signals to identify the remaining failure modes on top of this strong base capability. As a result, easy cases preserve broad coverage and stabilize common behaviors, while medium and hard cases contribute roughly $2{,}730$ examples with at least one verifier-detected failure that can further improve the model's upper-bound performance through SFT and verifier-derived RL. These failures are concentrated in \texttt{counting} ($11.6\%$), \texttt{lexical} ($10.2\%$), and \texttt{format} ($8.1\%$), whereas \texttt{style} ($0.16\%$) and \texttt{conditional} ($0\%$) are almost always satisfied. This gives later synthesis rounds a clear signal about which constraint types remain difficult, without sacrificing broad coverage or verifier reliability. While VISA introduces additional synthesis-time computation compared with static pipelines, this cost is incurred offline during data construction and is partly amortized by parallel candidate generation and batched verification; we discuss this trade-off in Appendix~\ref{app:efficiency}.

    \section{Conclusion}
        
     We presented \textsc{VISA}, an agentic framework for self-evolving multimodal instruction data synthesis. By coupling perception, planning and execution, reflection, and memory update, VISA turns static generate-and-filter pipelines into adaptive feedback-driven synthesis. Its persistent memory accumulates verifier outcomes, accepted samples, diversity signals, and target-model difficulty profiles, allowing later rounds to expand the constraint space and focus on unresolved weaknesses. 
    
    Experiments with Qwen3.5 target models demonstrate that \textsc{VISA} improves multimodal instruction following while preserving broad multimodal capabilities. \textsc{VISA}-SFT-15k surpasses both the base model and data-synthesis baselines on the average MM-IFEval score, while \textsc{VISA}-RL-15k further improves the average MM-IFEval score to 64.9. Across seven general multimodal benchmarks, \textsc{VISA}-RL-15k improves the average score from 70.5 to 72.9, showing that stronger instruction adherence does not come at the cost of general perception and reasoning ability. The same verifier contracts also serve directly as RL reward signals, requiring no separately trained reward model. We hope \textsc{VISA} provides a useful step toward adaptive, feedback-driven data synthesis for multimodal instruction-following models.

	\section*{Limitations}
	
	VISA primarily focuses on improving multimodal instruction synthesis through agentic iteration. As a result, the current framework prioritizes synthesis quality and difficulty adaptation over aggressive efficiency optimization. Although most synthesis stages can be parallelized across samples and constraints, state updates, iterative recovery, and target-aware probing still introduce additional overhead compared with static one-shot pipelines. We view this trade-off as acceptable in offline data construction, where synthesis quality is the primary objective.

	\section*{Ethics Statement}
	
	This work studies synthetic data generation for multimodal instruction following. The generated data may inherit biases, unsafe associations, or privacy-sensitive content from the source images and the MLLM used for synthesis. Although \ours{} uses image-aware constraint selection, verification, and reflection feedback to improve constraint satisfaction, these mechanisms are not complete safety or fairness filters. \ours{} is designed for research on multimodal instruction tuning and agentic data synthesis; models trained with its data should be audited before deployment, especially in high-stakes settings, and all source images and generated artifacts should be used in accordance with the licenses of the underlying datasets.
	
	\bibliography{custom}
	
	\appendix
	
	\section{Additional Analysis}
	\label{app:additional_analysis}
	
\begin{figure*}[t]
		\centering
		\includegraphics[width=0.58\textwidth]{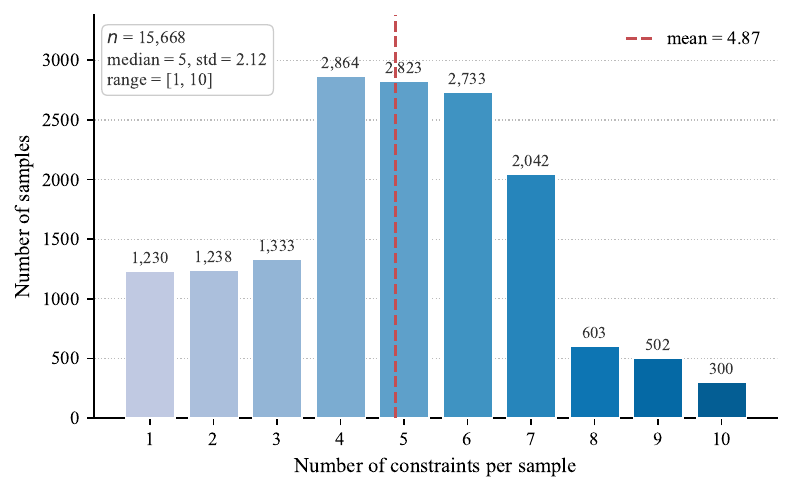}
		\caption{Distribution of the number of constraints per synthesized sample, showing that most accepted samples combine multiple jointly satisfiable requirements rather than a single constraint.}
		\label{fig:constraint_count}
	\end{figure*}

	Figure~\ref{fig:constraint_count} shows the full distribution of the number of constraints attached to each synthesized sample.
	
	\section{Constraint Details}
	\label{app:constraint_details}
	
    The constraint registry is organized at two granularities. Constraint categories provide coarse groups for analysis and diversity control, while constraint types are the atomic units sampled by the planner and bound to verifiers. Table~\ref{tab:constraint_details} summarizes the final registry after self-evolution. ``Seed'' counts the hand-crafted types available at initialization, and ``+Disc.'' counts types added through dynamic discovery after verifier binding. The largest expansions appear in visually grounded and structural categories, reflecting VISA's emphasis on image-compatible requirements and response organization.
    
    The seed constraint registry is initialized with common instruction-following constraint families studied in prior IF benchmarks, including IFEval and MM-IFEngine/MM-IFEval. VISA extends this initial registry through image-conditioned discovery and verifier binding. Unlike a fixed benchmark taxonomy, VISA maintains the registry as an evolving synthesis-time object whose entries can be filtered, discovered, validated, and reused across self-evolution rounds.
	
	\begin{table*}[t]
		\centering
		\small
		\setlength{\tabcolsep}{4pt}
		\renewcommand{\arraystretch}{1.15}
		\begin{tabular}{lp{10.0cm}ccc}
			\toprule
			\textbf{Category} & \textbf{Representative constraint types} & \textbf{Seed} & \textbf{+Disc.} & \textbf{Total} \\
			\midrule
			format     & \texttt{word\_count}, \texttt{sentence\_count}, \texttt{paragraph\_count}, \texttt{json\_format}, \texttt{xml\_format}, \texttt{yaml\_format}, \texttt{markdown\_format}, \texttt{list\_format}, \texttt{punctuation\_control}, \texttt{number\_precision}, \texttt{quotation}, \texttt{placeholder\_count}, \texttt{flexible\_layout} & 23 & +1 & 24 \\
			lexical    & \texttt{must\_include\_keyword}, \texttt{must\_exclude\_keyword}, \texttt{keyword\_frequency}, \texttt{keyword\_in\_first\_sentence}, \texttt{alliteration}, \texttt{acronym\_constraint}, \texttt{word\_diversity}, \texttt{synonym\_substitution} & 16 & +0 & 16 \\
			structural & \texttt{section\_headers}, \texttt{prefix\_suffix}, \texttt{paragraph\_start}, \texttt{paragraph\_end}, \texttt{sentence\_pattern}, \texttt{logical\_connector}, \texttt{summary}, \texttt{argument\_structure}, \texttt{postscript} & 13 & +53 & 66 \\
			style      & \texttt{tone}, \texttt{perspective}, \texttt{language}, \texttt{tense}, \texttt{rhetorical\_device}, \texttt{audience\_adaptation}, \texttt{role}, \texttt{scenario\_framing} & 11 & +2 & 13 \\
			\midrule
			attribute       & \texttt{attribute\_describe}, \texttt{scene\_mood}, \texttt{size\_compare}, \texttt{visual\_include}, \texttt{visual\_exclude} & 5  & +105 & 110 \\
			spatial         & \texttt{spatial\_relation} & 1  & +70 & 71 \\
			counting        & \texttt{object\_count} & 1  & +1 & 2 \\
			object\_grounded & \texttt{object\_describe} & 1  & +0 & 1 \\
			ocr\_grounded    & \texttt{ocr\_content} & 1  & +1 & 2 \\
			causal          & \texttt{causal\_explain}, \texttt{temporal\_sequence} & 2  & +1 & 3 \\
			conditional     & \texttt{conditional\_if}, \texttt{hypothetical} & 2  & +0 & 2 \\
			\midrule
			\multicolumn{2}{r}{\textbf{Total constraint types}} & \textbf{76} & \textbf{+234} & \textbf{310} \\
			\bottomrule
		\end{tabular}
		\caption{Constraint categories in the final VISA registry. Representative types are illustrative examples; “Seed” and “+Disc.” denote initially defined and dynamically discovered constraint types, respectively.}
		\label{tab:constraint_details}
	\end{table*}

	\section{Discussion on Synthesis Efficiency}
	\label{app:efficiency}
	
	The main cost of \ours{} comes from replacing single-pass generation with an agentic loop. In our implementation, the planner generates four candidate instructions for each image during the greedy selection stage, and we set the multi-threaded concurrency to 16. With a single synthesis model deployed on L40s GPUs, the end-to-end synthesis time is approximately 20 seconds per accepted sample on average.
	
	Several design choices help amortize this cost. Perception is performed once per image; the four candidate instructions are generated in parallel before the more expensive response-generation stage; code-based verifiers are lightweight and parallelizable; and LLM-track constraints are packed into batched judge calls whenever possible. The bounded recovery mechanism also avoids unbounded self-correction. Among the 15,668 accepted samples, 10,230 samples (65.3\%) pass verification without recovery, 2,241 samples (14.3\%) require one recovery attempt, 2,174 samples (13.9\%) require two attempts, and 1,023 samples (6.5\%) reach three attempts. This shows that most accepted samples incur no recovery overhead, while only a small fraction requires the maximum repair budget.
	
	The efficiency trade-off is therefore best understood per accepted and target-informative sample rather than per raw generated sample. A static pipeline may generate more candidates per unit time, but many candidates can be redundant, incompatible with the image, unverifiable, or already easy for the target model. In contrast, \ours{} spends more computation before admission, but each accepted sample is associated with explicit constraint sets, verifier handles, difficulty scores, and recovery diagnostics. These metadata are not discarded: they update the agent state and guide subsequent synthesis rounds. To assess the reliability of the LLM-based judge track, we further conduct a manual audit on a sampled subset of judge decisions and observe approximately 95\% agreement with human annotations, suggesting that the verifier feedback used by the agent is largely consistent with human judgment.
    
	\section{Tool Registry Interface}
	\label{app:tool}
	
	Each verification tool in \textsc{VISA}'s code-verifier track follows a compact interface contract: a metadata block links the tool to a constraint category and type, a \texttt{verify} function maps a response and typed parameters to a standardized pass/fail record, and a small set of test cases is executed before the tool enters the registry. The same contract is used for both seed verifiers and code verifiers attached to dynamically discovered constraint types. The schematic below illustrates the interface with a shortened JSON-format verifier; implementation details are omitted because they are tool-specific and not central to the framework.
	
	\begin{scriptsize}
		\begin{verbatim}
			TOOL_META = {
				"tool_id": "verify_json_format",
				"description": "Check JSON format",
				"constraint_categories": ["format"],
				"constraint_types": ["json_format"],
				"input_schema": {
					"response": {
						"type": "str",
						"required": True,
						"desc": "Response text"
					},
					"required_keys": {
						"type": "list",
						"required": False,
						"desc": "Required JSON keys"
					}
				},
				"output_schema": {
					"pass": {"type": "bool"},
					"expected": {"type": "str"},
					"actual": {"type": "str"},
					"detail": {"type": "str"}
				}
			}
			
			def verify(response: str, **params) -> dict:
			...
			return {
				"pass": <bool>,
				"expected": <str>,
				"actual": <str>,
				"detail": <str>
			}
			
			TEST_CASES = [
			{"input": {"response": '{"a": 1}'},
				"expected_pass": True},
			{"input": {"response": "not json"},
				"expected_pass": False},
			...
			]
		\end{verbatim}
	\end{scriptsize}

	\section{Prompt Examples}
	\label{app:prompts}
	
	This appendix provides abbreviated prompt templates for the main agent calls in \textsc{VISA}. We omit implementation-specific formatting instructions and long field lists; placeholders such as \texttt{\{\ldots\}} are filled by the controller at runtime.
	
	\paragraph{Perception and constraint-type discovery.}
	Issued at the beginning of every iteration to the multimodal teacher together with the input image. It returns both the perception profile of the image and, when warranted, proposals for new constraint types.
	
	\begin{quote}\small\itshape
		Analyze the image and return a compact structured profile. Summarize the visible scene, estimate its complexity, and mark any vision-grounded constraint types from \{registered types\} that the image cannot support.\\[2pt]
		If the registered library misses a useful requirement for this image-conditioned instruction, propose up to \{max\_discover\} new constraint types with their category, verifier track, and short description. Do not propose a new type when an existing parameterized type already covers the requirement.
	\end{quote}
	
	\paragraph{Constraint selection.}
	Given the perception profile and the candidate constraint types compatible with the image, the planner selects a budgeted, mutually consistent constraint set.
	
	\begin{quote}\small\itshape
		Select \{constraint\_budget\} compatible constraint types from \{candidate types\}. Use the priority weights from the agent state as a soft guide, while avoiding redundant choices from a single category.\\[2pt]
		The selected constraints must be jointly satisfiable and verifiable. For each selected type, write a short description: vision-grounded constraints should refer to objects by role or category without revealing the answer, and text-only constraints should state the required response property directly.
	\end{quote}
	
	\paragraph{Instruction composition.}
	Once a constraint set is fixed, the agent rewrites it as a single natural-language user instruction.
	
	\begin{quote}\small\itshape
		Write a single natural user request for the image. The request should be fluent and self-contained, and it should integrate all selected constraints without listing them mechanically.\\[2pt]
		Do not reveal visual evidence that the model should infer from the image. If a constraint specifies the response format, ask for that format in natural language rather than providing an answer skeleton.
	\end{quote}
	
	\paragraph{Verifier parameter extraction.}
	For constraints bound to a code verifier, a lightweight extraction prompt fills the verifier's typed parameters from the natural-language description.
	
	\begin{quote}\small\itshape
		For each code-verifiable constraint, extract only the parameters explicitly supported by its verifier schema. Preserve the required data types and return one parameter object per constraint.\\[2pt]
		If a value is not specified or not relevant, set it to \texttt{null}; do not infer missing values from common sense or from the image.
	\end{quote}
	
	\paragraph{Coverage self-check.}
	Before full verification, a judge LLM checks whether the composed instruction explicitly references every selected constraint, and feeds any missing items back to the recovery loop.
	
	\begin{quote}\small\itshape
		Check whether the composed instruction explicitly covers every selected constraint. For each missing constraint, return its identifier and a brief repair suggestion.\\[2pt]
		Do not count a constraint as covered merely because it is implicit, conventional, or likely to be satisfied by default.
	\end{quote}
	
\end{document}